\documentclass[letterpaper, 10 pt, conference]{ieeeconf}  % Comment this line out if you need a4paper

\IEEEoverridecommandlockouts                              % This command is only needed if 
\usepackage{cite}
\usepackage{amsmath}
\usepackage{cleveref}
\usepackage{amssymb}  % assumes amsmath package installed

\usepackage{booktabs,multirow}
\usepackage[table]{xcolor}
\usepackage{adjustbox} % for width=\textwidth wrapper

\definecolor{best}{RGB}{255,199,206}
\definecolor{second}{RGB}{255,235,156}
\newcommand{\best}[1]{\cellcolor{best}\textbf{#1}}
\newcommand{\second}[1]{\cellcolor{second}#1}

\title{\LARGE \bf
Nighttime Autonomous Driving Scene Reconstruction \\ with Physically-Based Gaussian Splatting
}

\author{Tae-Kyeong Kim$^{1,2}$, Xingxin Chen$^{1}$, Guile Wu$^{1}$, Chengjie Huang$^{1}$, Dongfeng Bai$^{1}$, and Bingbing Liu$^{1}$ 
\thanks{$^{1}$The authors are with Huawei Noah's Ark Lab.
$^{2}$Tae-Kyeong Kim is with the University of Toronto and contributed to this work during an internship at Huawei Noah's Ark Lab, Canada.
taekyeong.kim@mail.utoronto.ca.
{\{xingxin.chen, baidongfeng, liu.bingbing\}@huawei.com},
{\{guile.wu, chengjie.huang\}@outlook.com}.%
}
}

\begin{document}

\maketitle
\thispagestyle{empty}
\pagestyle{empty}

%%%%%%%%%%%%%%%%%%%%%%%%%%%%%%%%%%%%%%%%%%%%%%%%%%%%%%%%%%%%%%%%%%%%%%%%%%%%%%%%
\begin{abstract}
This paper focuses on scene reconstruction under nighttime conditions in autonomous driving simulation.
Recent methods based on Neural Radiance Fields (NeRFs) and 3D Gaussian Splatting (3DGS) have achieved photorealistic modeling in autonomous driving scene reconstruction, but they primarily focus on normal-light conditions.
Low-light driving scenes are more challenging to model due to their complex lighting and appearance conditions, which often causes performance degradation of existing methods.
To address this problem, this work presents a novel approach that integrates physically based rendering into 3DGS to enhance nighttime scene reconstruction for autonomous driving.
Specifically, our approach integrates physically based rendering into composite scene Gaussian representations and jointly optimizes Bidirectional Reflectance Distribution Function (BRDF) based material properties.
We explicitly model diffuse components through a global illumination module and specular components by anisotropic spherical Gaussians.
As a result, our approach improves reconstruction quality for outdoor nighttime driving scenes, while maintaining real-time rendering.
Extensive experiments across diverse nighttime scenarios on two real-world autonomous driving datasets, including nuScenes and Waymo, demonstrate that our approach outperforms the state-of-the-art methods both quantitatively and qualitatively.
\end{abstract}

%%%%%%%%%%%%%%%%%%%%%%%%%%%%%%%%%%%%%%%%%%%%%%%%%%%%%%%%%%%%%%%%%%%%%%%%%%%%%%%%
\section{INTRODUCTION}

With the rapid development of autonomous driving, the demand for safety continues to increase.
Creating digital twins of driving scenes has gained increasing attention because these digital twins can be used to simulate safety-critical corner cases that are difficult and costly to capture in the real world.
By performing closed-loop evaluation with these reconstructed driving scenes, the reliability of autonomous driving systems can be continuously improved.

Contemporary driving scene reconstruction methods can be categorized into Neural Radiance Fields (NeRF) based methods and 3D Gaussian Splatting (3DGS) based methods.
NeRF-based methods implicitly model driving scenes with neural scene graphs and multi-layer perceptrons (MLPs).
Although NeRF-based methods have achieved photorealistic scene modeling~\cite{ost2021neural,cao2024lightning,yang2023unisim, yang2023emernerf}, they inherently suffer from slow training and rendering speeds.
On the other hand, 3DGS-based methods explicitly employ composite Gaussian primitives and tile-based differentiable rasterization for driving scene reconstruction~\cite{chen2025omnire,yan2024street,wu2025armgs,zhou2024drivinggaussian}.
They have shown promising results for modeling dynamic driving scenes, while maintaining real-time rendering performance.
However, existing methods primarily focus on driving scene reconstruction under normal-light conditions, neglecting more challenging nighttime driving scene reconstruction.
Consequently, existing methods often suffer from performance degradation for nighttime scene reconstruction due to complex lighting and appearance conditions under low-light conditions.
Although scene reconstruction under low-light conditions~\cite{qu2024lush,ye2024gaussian,gao2024relightable} has been explored in some other fields, these methods are mostly designed for static scenes and cannot be directly used for dynamic driving scene modeling.
In addition, their reliance on environment maps may lead to failures in urban driving scenes due to complex lighting and appearance conditions.

To address these problems, this work presents a novel approach that integrates physically based rendering into composite scene Gaussian representations for autonomous driving scene reconstruction.
Specifically, our approach disentangles diffuse and specular component modeling and jointly optimizes Bidirectional Reflectance Distribution Function (BRDF) based material properties to enhance nighttime scene reconstruction.
Unlike prior work that relies on environment map sampling, we propose a global illumination module to model the diffuse component.
For specular component, we employ anisotropic spherical Gaussians (ASGs)~\cite{xu2013anisotropic} to better capture high-frequency specular effects and equip each Gaussian with BRDF-based material properties.
Then, we solve a physically based rendering function to obtain color contribution of each Gaussian and map High Dynamic Range (HDR) colors to Low Dynamic Range (LDR) colors before differentiable rasterization.
In this way, our approach is capable of improving outdoor nighttime driving scene reconstruction while maintaining real-time rendering. 
Our experiments on two real-world autonomous driving datasets, including nuScenes~\cite{caesar2020nuscenes} and Waymo~\cite{sun2020scalability}, demonstrate that our approach achieves the state-of-the-art performance compared with existing methods for nighttime driving scene reconstruction.
In summary, our contributions are:
\begin{itemize}
    \item We propose a novel method to model nighttime driving scene by integrating physically based rendering into composite scene Gaussian representations, which fills a gap left by existing works.
    \item We design a global illumination module to model diffuse component which does not require per-ray environment map sampling and is suitable for dynamic scene modeling.
    \item We combine ASGs with BRDF-constrained rendering for specular component modeling to capture high-frequency effects while ensuring physical plausibility.
\end{itemize}

\section{Related Works}

\subsection{Gaussian Splatting for Driving Scene Reconstruction}
Recently, Gaussian splatting has been widely used for driving scene reconstruction in autonomous driving simulation due to its faster training and real-time rendering efficiency.
To model dynamic driving scenes, researchers often employ composite scene Gaussian representations, which decompose scenes into dynamic object nodes, static background nodes and distant region nodes in Gaussian scene graphs~\cite{chen2025omnire,yan2024street,wu2025armgs,zhou2024drivinggaussian,chen2023periodic,ren2024unigaussian}.
OmniRe~\cite{chen2025omnire} constructs a Gaussian scene graph of diverse Gaussian node types to unify reconstruction of rigid vehicles, deformable agents, and non-rigid backgrounds, which achieves state-of-the-art dynamic reconstruction for autonomous driving.
StreetGS~\cite{yan2024street} introduces decomposition of urban environments into background, foreground actors, and sky, using tracked poses and cubemaps for large-scale rendering.
HUGS~\cite{zhou2024hugs} and ArmGS~\cite{wu2025armgs} integrate appearance modeling into composite scene Gaussian representations to model camera exposure and dynamic scene conditions.
UniGaussian~\cite{ren2024unigaussian} introduces multi-modal multi-sensor simulation using composite scene Gaussian representations.
However, these works primarily focus on driving scene reconstruction under normal-light conditions, leaving low-light (nighttime) scenarios largely underexplored.
This motivates us to develop a novel approach that remains effective under challenging nighttime conditions.
To this end, our work adopts a Gaussian scene graph for dynamic driving scene reconstruction and integrates physically based rendering into composite scene Gaussian representations, which enables effective modeling of complex illumination effects in nighttime environments.

\begin{figure*}[t]
  \centering
  \includegraphics[width=\textwidth,keepaspectratio]{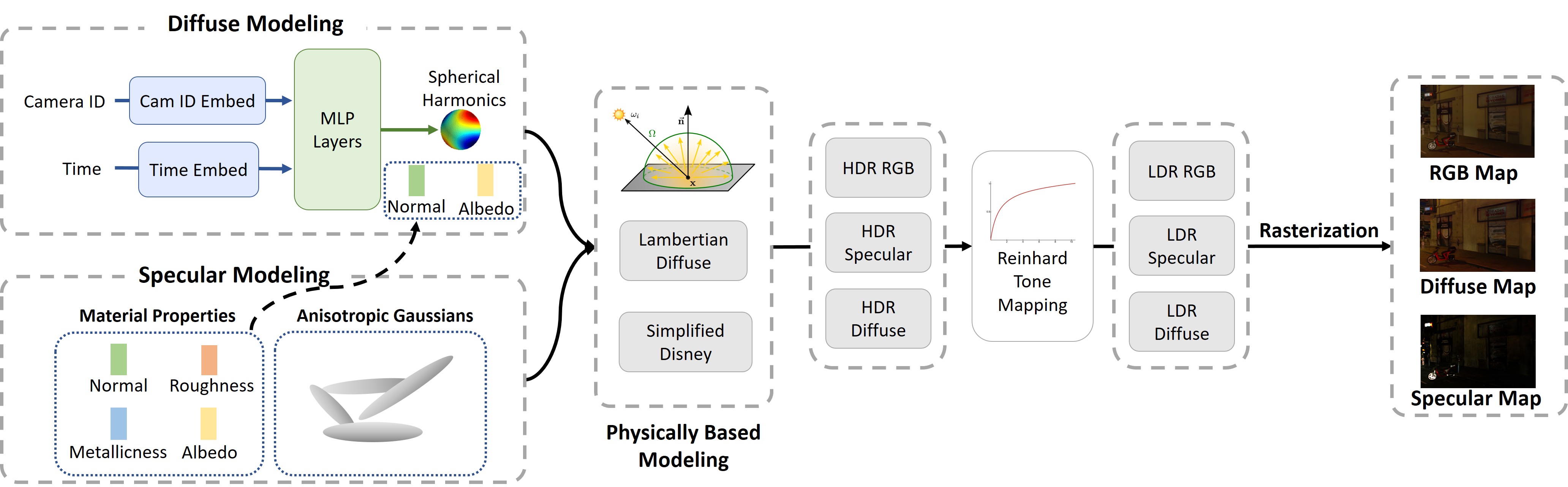}
  \caption{An overview of our approach. Our framework decomposes lighting into specular and diffuse component. Each of the respective component is modeled using the per-Gaussian ASGs and global SH module that are constrained by the BRDF}
  \label{fig:model}
\end{figure*}

\subsection{Lighting Modeling for Scene Reconstruction}
Lighting modeling for scene reconstruction has been actively explored to improve reconstruction quality under complex illumination.
One line of work incorporates physically based rendering (PBR), which is widely used in computer graphics for physically plausible relighting.
NeILF~\cite{yao2022neilf} first introduces per-scene material parameters and enables relighting via the Simplified Disney BRDF model~\cite{burley2012physically}.
Building on this idea, RelightableGS~\cite{gao2024relightable} extends PBR to 3DGS, assigning material attributes (albedo, roughness, metallic) to each Gaussian, and use a point-based rasterizer to produce the final image from the shaded Gaussians.
However, these relighting methods are designed mainly for static scenes.
They usually require per-ray sampling of environment maps or incident lighting at each rendering step, which slows down inference compared to standard SH-based 3DGS.
In addition, they often rely on extended shading pipelines rather than the lightweight rasterization used in vanilla 3DGS. 
More recently, Spec-Gaussian~\cite{yang2024spec} employs per-Gaussian ASGs for modeling sharper highlights.
DarkGS~\cite{zhang2024darkgs} and LL-Gaussian~\cite{sun2025ll} enhance the robustness of 3DGS in low-light scenarios.
However, these methods still focus on static scenes and cannot be directly used for dynamic driving scene modeling.
Different from these works, we integrate physically based rendering into composite scene Gaussian representations and present a global illumination module to model diffuse component and BRDF-constrained rendering with ASGs to model specular component for autonomous driving simulation.

\section{Methods}
\subsection{Preliminaries}
\subsubsection{Gaussian Splatting}
3D Gaussian Splatting (3DGS) \cite{kerbl20233d} represents a scene as a collection of colorized 3D Gaussian primitives. Each Gaussian is parameterized by opacity $o \in [0,1]$, position $\mu \in \mathbb{R}^3$, covariance matrix $\Sigma$ (derived from a quaternion rotation $q \in \mathbb{R}^4$ and scaling vector $s \in \mathbb{R}^3$), and color $c \in \mathbb{R}^F$, often expressed as spherical harmonics (SH) coefficients.
These parameters define the Gaussian in 3D space as the following:\begin{equation}
G(x) = \exp\left(-\tfrac{1}{2}(x-\mu)^{T}\Sigma^{-1}(x-\mu)\right).
\label{3DGS}
\end{equation}
During rendering stage, the 3D Gaussians are projected onto the image plane as 2D Gaussians. The color of each pixel is then computed by alpha-blending N-ordered 2D Gaussians:
\begin{equation}
C = \sum_{i \in N} c_i \alpha_i \prod_{j=1}^{i-1} (1 - \alpha_j),
\label{color blending}
\end{equation}
where $c_i$ is the color derived from SH coefficients and $\alpha_i$
is determined by the opacity of the Gaussian.

\subsubsection{Composite Scene Graph Structure}
Following previous work~\cite{chen2025omnire,yan2024street,wu2025armgs,zhou2024drivinggaussian,chen2023periodic,ren2024unigaussian}, we construct a 3DGS scene graph that consists of three components: (i) static background, (ii) dynamic actors, and (iii) sky map. Background Gaussians are initialized from LiDAR points. Rigid objects (e.g., vehicles) are represented by Gaussian nodes whose positions are updated according to the vehicle trajectory $T_v \in \mathrm{SE}(3)$. Non-rigid objects are modeled as SMPL Gaussians for pedestrians~\cite{chen2025omnire} and deformable Gaussians for cyclists and distant pedestrians~\cite{yang2024deformable}. The sky region is modeled using an explicit cubemap representation, which is sampled on given viewing direction.
While this scene structure provides a robust representation of autonomous driving scenes, our method further integrates a lighting module to achieve photorealistic reconstructions under challenging low-light conditions.

\subsection{Approach Overview}
An overview of our proposed framework for nighttime autonomous driving scene reconstruction is shown in \Cref{fig:model}.
Our method reconstructs dynamic nighttime driving scenes by integrating physically based rendering with a composite 3DGS scene graph. Specifically, for a given camera and timestep, we decompose the Gaussian appearance attributes into diffuse and specular components: a global lighting module predicts SH coefficients to model the diffuse illumination, while per-Gaussian ASGs capture the incident specular lighting. The two components are synergically combined with additional material attributes to produce HDR radiance, which is tone-mapped to LDR via Reinhard and rendered with the standard differentiable 3DGS rasterizer.

\subsection{BRDF Modeling}
Our lighting method utilizes the physically based rendering equation~\cite{kajiya1986rendering} to compute the interaction between the lighting of the scene and the Gaussian surfaces. The rendering equation is defined as follows:
\begin{equation}
     L_{o}(\boldsymbol {\omega _{o}}, \boldsymbol {x}) = \int _{\Omega }f(\boldsymbol {\omega _{o}}, \boldsymbol {\omega _{i}}, \boldsymbol {x})L_{i}(\boldsymbol {\omega _{i}}, \boldsymbol {x})(\boldsymbol {\omega _{i}}\cdot \boldsymbol {n})d\boldsymbol {\omega _{i}}, \label {eq:rendering_equation}
\end{equation}
where $x$ denotes a surface point with normal vector $n$, $f$ is the BRDF function, and $L_i$ and $L_o$ represent the incoming and outgoing radiance in directions $\omega_i$ and $\omega_o$, respectively. The domain $\Omega$ corresponds to the hemisphere above the surface.
Following~\cite{gao2024relightable}, each Gaussian primitive is parameterized as $g(\mu,q,s; b,r,m,n)$, where in addition to the mean $\mu$ and covariance (represented by rotation quaternion $q$ and scale $s$), each Gaussian primitive is assigned material attributes and a surface normal:
\begin{itemize}
    \item $b \in [0,1]^3$: albedo (base color),
    \item $r \in [0,1]$: roughness,
    \item $m \in [0,1]$: metallicness,
    \item $n \in \mathbb{R}^3$: surface normal.
\end{itemize}

\subsection{Global Lighting Module}
Previous work~\cite{gao2024relightable,yao2022neilf} models global incident light with environment maps, which requires sampling for every rendering step. Additionally, they are primarily tested for static scenes, and building a compact robust environment map for dynamic scenes is challenging.
To address these limitations, we propose a global lighting module that predicts scene illumination conditioned on normalized timestep and camera embeddings.
Specifically, we use an MLP to produce a latent representation, which is passed through a detection head for each SH level.
In our design, we employ second-degree SH, producing three sets of coefficients.
These SH coefficients are combined with the normal $n$ and albedo $b$ properties of the Gaussians, which are then evaluated using the Lambertian diffuse equation \cite{cohen1993radiosity}. Following~\cite{green2003spherical}, the diffuse contribution can be approximated as a weighted summation over SH coefficients:
\begin{equation}
    L_d = \frac{b}{\pi}\sum_{l=0}^{2}A_{l}\sum_{m=-l}^{l}c^{l}_{m}Y^{l}_{m}(n),
    \label{Diffuse}
\end{equation}
where $A_l$ is the cosine-kernel convolution factors that convert incident radiance SH into irradiance SH for a Lambertian, $Y_{l}^{m}$ represents the real SH basis evaluated at Gaussian normal, and $c_{l}^{m}$ is the SH coefficients produced by the environment lighting module.

\subsection{Specular Light Modeling}
Nighttime driving scenes often contain artificial lighting and intense reflections from retro reflectors. To better handle these situations, we employ ASGs~\cite{xu2013anisotropic} within our rendering pipeline, as they have been shown to capture specular highlights more effectively than SH-based approaches~\cite{gao2024relightable}.
Following~\cite{xu2013anisotropic}, an ASG is defined as:
\begin{equation}
    G(v; [x, y, z], [\lambda, \mu], c) =
    c \cdot S(v; z) \cdot
    e^{-\lambda (v \cdot x)^2 - \mu (v \cdot y)^2},
\end{equation}
where $x, y, z$ represents the lobe direction, $\lambda, \mu$ denotes the the lobe sharpness in the respective $x$ and $y$ coordinates, and $S$ is the saturation between the viewing direction $v$ and the lobe $z$ axis.
ASGs can capture specular highlights more effectively than previous approaches based on SH coefficients to approximate incident specular lighting.
Instead of using ASGs as a latent representation passed to an MLP for the final RGB contribution like~\cite{yang2024spec},
our framework uses ASGs to model incident specular lighting. Each Gaussian is assigned a small set of ASG lobes (four in our case), and their specular contribution is evaluated during rendering via a BRDF-constrained PBR.
Following~\cite{yao2022neilf, gao2024relightable}, we use a simplified Disney BRDF representation:
\begin{equation}
    f_s(w_o, w_i) = \frac{D(h;r) \cdot F(w_o, h; b, m) \cdot G(w_i, w_o, h; r)}{(n \cdot w_i) \cdot (n \cdot w_o)},
    \label{Disney BRDF}
\end{equation}
where the $h$ represents the half vector between incoming and outcoming radiance, and $D$, $F$ and $G$ indicate the normal distribution function, the Fresnel term, and the geometry term respectively.
As the normal distribution can be modeled as a spherical Gaussian (SG)~\cite{gao2024relightable, yao2022neilf}, the convolution of the SG and ASG product inside the PBR can be simplified~\cite{xu2013anisotropic}:
\begin{equation}
\begin{split}
L_s &= G(\omega_i,\omega_0,h; r) \cdot F(\omega_o,h; b,m) \cdot \sum_{i} A_i \cdot s_i \cdot 
    \\ &ASG_i\!\left(w_r,[x_i,y_i,z_i], 
    \left[\tfrac{\nu\lambda_i}{\nu+\lambda_i}, \tfrac{\nu\mu_i}{\nu+\mu_i}\right],\,
    a_{\text{ndf}} \tfrac{\pi}{\sqrt{(\nu+\lambda_i)(\nu+\mu_i)}}\right),
\end{split}
\label{Simplified}
\end{equation}
where $\nu$ and $a_{ndf}$ represent concentration parameter and amplitude of the normal distribution, and $w_r$ denotes the reflection direction between incoming radiance and the Gaussian normal.
With this simplification, our approach can efficiently apply specular lighting to every Gaussian without any additional sampling. 

\subsection{Final Lighting and Tone Mapping}
The diffuse and specular lighting is summed together to produce the final relighted value for each Gaussian in HDR:
\begin{equation}
    L_{\text{HDR}} = L_{d} + L_{s}.
\end{equation}
Since the image pixel values are expressed in LDR values, we apply Reinhard tone mapping~\cite{reinhard2023photographic} to produce LDR values:
\begin{equation}
    L_{\text{LDR}} = \frac{L_{\text{HDR}}}{1 + L_{\text{HDR}}}.
\end{equation}
The resulting \(L_{\text{LDR}}\) serves as the per-Gaussian color \(c_i\) used for the standard rasterization.

\subsection{Optimization}

In order to obtain realistic lighting effects, accurate surface normals are crucial for each Gaussian. We employ an off-the-shelf model~\cite{wang2025moge} to obtain a normal map prior $N$ for the input image. During rasterization, the per-Gaussian normals $n_i$ are accumulated to produce the rendered normal map:
\begin{equation}
\hat{N} =
\frac{
\sum_{i \in M} n_i \alpha_i \prod_{j=1}^{i-1} (1 - \alpha_j)
}{
\sum_{i \in M} \alpha_i \prod_{j=1}^{i-1} (1 - \alpha_j)
},
\label{eq:weighted_normal}
\end{equation}
where $\alpha_i$ denotes the opacity contribution of the $i$-th Gaussian along the ray, and $M$ is the set of overlapping Gaussians.
To improve normal supervision with these priors, we follow the depth-normal loss regularizer from~\cite{chen2024vcr}:
\begin{equation}
L_n = \lVert \hat{N} - N \rVert_1 + \big(1 - \hat{N} \cdot N \big),
\label{eq:normal_loss}
\end{equation}

Additionally, we derive depth-based normals $\hat{N}_d$ from the rendered depth map and compute a confidence weight to measure the consistency between $\hat{N}_d$ and the prior normal $N$:

\begin{equation}
w = \exp \left( \frac{\hat{N}_d \cdot N - 1}{\gamma} \right),
\label{eq:confidence}
\end{equation}
where $\gamma$ is a hyperparameter controlling the sharpness of the weighting. Using this, we define the view-consistent D-Normal regularizer~\cite{chen2024vcr} as:
\begin{equation}
L_{dn} = w \cdot \left( \lVert \hat{N}_d - N \rVert_1 + \big( 1 - \hat{N}_d \cdot N \big) \right).
\label{eq:view_consistent_dn}
\end{equation}

The overall training objective combines the depth-normal loss with the standard RGB pixel and D-SSIM loss~\cite{chen2025omnire, yan2024street, zhou2024hugs}, which can formulated as:
\begin{equation}
L_{total} = w_{\text{rgb}} \cdot L_{\text{rgb}} + w_{\text{D-SSIM}} \cdot L_{\text{D-SSIM}} +  w_{dn} \cdot (L_d + L_n).
\label{eq:total_loss}
\end{equation}

\begin{figure*}[!t]
  \centering
  \includegraphics[width=0.95\textwidth,height=0.9\textheight,keepaspectratio]{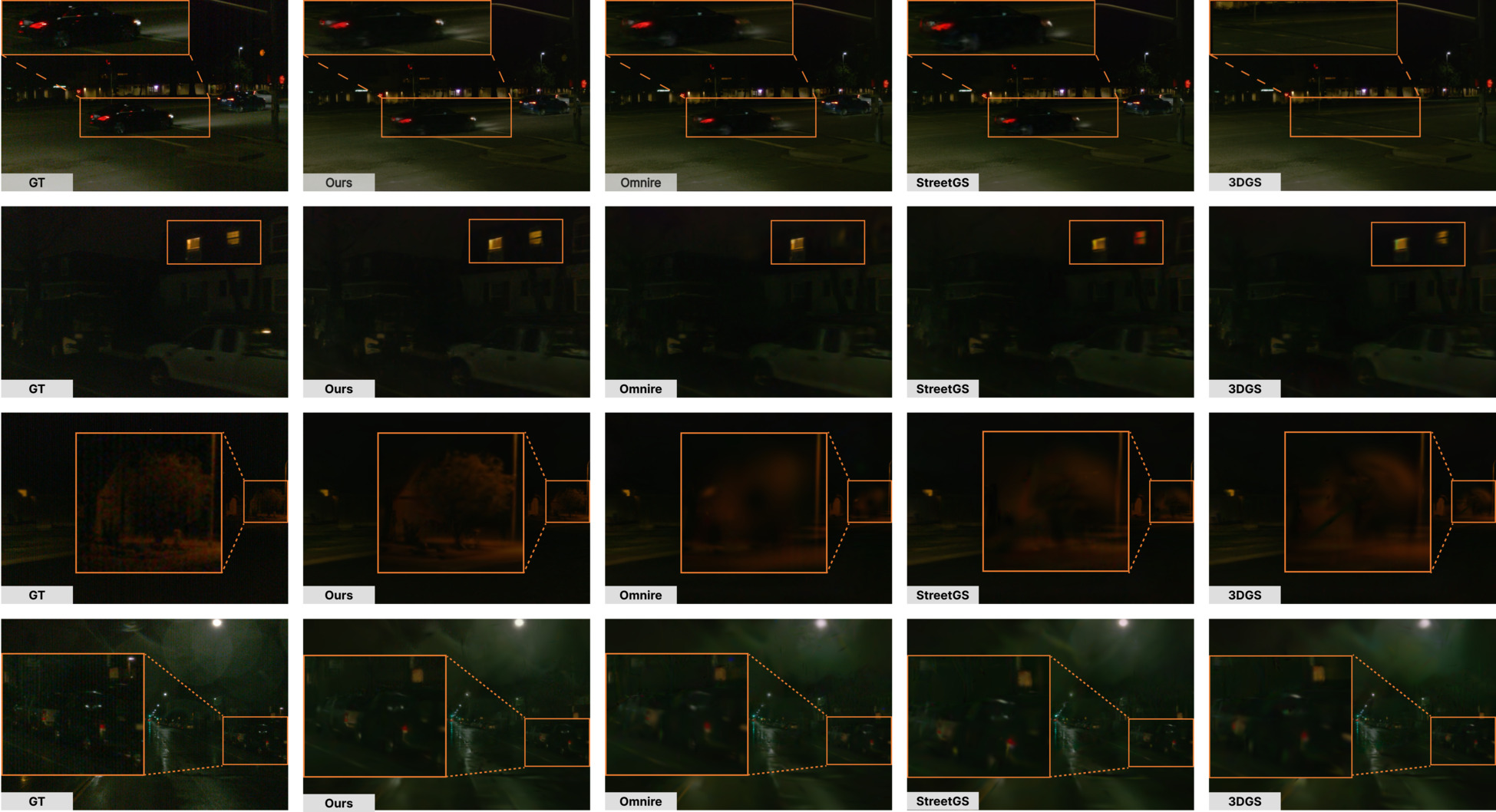}
  \caption{Qualitative results from the Waymo Open Dataset. Columns show results from GT, Ours, OmniRe, StreetGS, and 3DGS, respectively. Fine-grained details such as traffic lights, vehicles, and trees are highlighted to illustrate differences in reconstruction quality.}
  \label{fig:waymo}
\end{figure*}

\section{Experiment}

\subsection{Datasets}
We conducted our experiments on two challenging real-world autonomous driving datasets, including nuScenes~\cite{caesar2020nuscenes} and the Waymo Open Dataset~\cite{sun2020scalability}.
The Waymo Open Dataset contains high-resolution images with challenging nighttime scenes, while the nuScenes dataset has lower resolution but provides diverse urban scenes with low ambient lighting and sharp highlights.
In this work, we consider \emph{low-light dynamic scenes} as those that exhibit low ambient illumination from the sky and contain sharp localized lighting sources (e.g., headlights, buildings, street lamps), which reflects real-world challenges for autonomous driving in low-light conditions.
We therefore select 11 low-light driving scenes from each datasets for our experiments.
Specifically, we use scenes 754, 763, 774, 776, 780, 781, 782, 783, 784, 785, and 790 for nuScenes,
and scenes 007, 012, 015, 018, 030, 038, 051, 099, 106, 129, and 166 for Waymo.
For evaluation, every eighth frame is reserved for novel view testing, while the remaining frames are used for training.

\subsection{Implementation Details}
In our experiments, each driving scene is initialized with LiDAR point clouds, supplemented by uniformly sampled points following~\cite{chen2023periodic}. The model is trained for 40,000 iterations using the Adam optimizer~\cite{kingma2014adam}. The learning rate for normals, roughness, and metallicness is set to $1\times 10^{-3}$, while the learning rate for Gaussian lobe axes, stretch parameters along the $x$ and $y$ coordinates, and amplitude is set to $1\times 10^{-5}$. The global SH illumination module consists of 8 linear layers, and each latent representation is passed through a single linear head to produce the SH coefficients at each level. For training, the loss weights are empirically set as $\omega_\text{rgb}=0.8$, $\omega_\text{D-SSIM}=0.2$, and $\omega_\text{LPIPS}=0.025$.

\subsection{Baselines}
We compare our framework against recent state-of-the-art 3DGS methods for dynamic urban scene reconstruction: OmniRe~\cite{chen2025omnire} and StreetGaussians~\cite{yan2024street}, We also include 3DGS~\cite{kerbl20233d} as a baseline for comparison.
Note that, since previous work~\cite{chen2025omnire,yan2024street} has demonstrated the superiority of 3DGS-based method over NeRF-based methods, we do not include comparison with NeRF-based methods in this work.
For evaluation metrics, following~\cite{chen2025omnire,yan2024street}, we evaluate PSNR for pixel-level, SSIM for structural-level~\cite{zhang2018unreasonable}, and LPIPS~\cite{wang2004image} for perceptual similarity.

\begin{table}[t]
\caption{Quantitative comparison on the Waymo dataset for scene reconstruction and novel view synthesis tasks.}
\label{tab:waymo}
\centering
\scriptsize
\setlength{\tabcolsep}{4pt}
\renewcommand{\arraystretch}{1.12}
\begin{adjustbox}{width=0.48\textwidth} % half the column width
\begin{tabular}{lccc ccc}
\toprule
\multirow{2}{*}{Method} &
\multicolumn{3}{c}{Reconstruction} &
\multicolumn{3}{c}{Novel View} \\
\cmidrule(lr){2-4}\cmidrule(lr){5-7}
& PSNR$\uparrow$ & SSIM$\uparrow$ & LPIPS$\downarrow$
& PSNR$\uparrow$ & SSIM$\uparrow$ & LPIPS$\downarrow$ \\
\midrule
3DGS~\cite{kerbl20233d} 
& 30.1 & 0.755 & 0.487
& 28.3 & 0.720 & 0.503 \\

StreetGaussians~\cite{yan2024street} 
& 30.5 & 0.757 & 0.482
& 28.5 & \best{0.722} & 0.499 \\

OmniRe~\cite{chen2025omnire}
& \second{31.0} & \second{0.768} & \second{0.455}
& 28.5 & 0.718 & \second{0.481} \\

Ours          
& \best{31.9} & \best{0.781} & \best{0.441}
& \best{28.8} & \second{0.720} & \best{0.467} \\
\bottomrule
\end{tabular}
\end{adjustbox}
\end{table}

\subsection{Results on Waymo}
The quantitative results on Waymo are reported in~\Cref{tab:waymo}.
Overall, our approach consistently outperforms prior state-of-the-art methods on both novel view synthesis and scene reconstruction. 
Specifically, for scene reconstruction, our model achieves a PSNR of 31.9 dB, SSIM of 0.781, and LPIPS of 0.441, surpassing any other existing models. We attribute these improvements to our decomposed lighting module: SH-based global illumination enhances stability across viewpoints and timestamps, while per-Gaussian ASGs effectively capture localized highlights such as headlights and reflective surfaces.
For novel view synthesis, our model achieves the highest overall perceptual quality, with a PSNR of 28.8 dB, SSIM of 0.720, and LPIPS of 0.467. Although the PSNR and SSIM improvements are moderate, the consistent reduction in LPIPS underscores the ability of our lighting decomposition to produce sharper and more visually realistic renderings under low-light conditions. Importantly, despite the inherent challenge of low-light driving scenarios, our method demonstrates consistent performance across 11 diverse scenes.

A qualitative comparison on Waymo nighttime sequences is shown in~\Cref{fig:waymo}.
We can see that our method produces clearer vehicle renderings and accurately reconstructs the effect of headlights on the road surface. Additionally, it recovers more fine-grained details, such as illuminated building windows and reflective surfaces, which other baselines tend to oversmooth or fail to capture. For instance, vehicles rendered by other methods are overly smoothed and fail to model the car headlights and their interaction with the road, while our method generates sharper vehicles and captures localized lighting effects on the background. These results highlight the superiority of our framework in modeling low-light driving scenes. Furthermore, our approach also reconstructs clearer obstacles, such as trees, even under extremely low-light conditions, which is crucial for autonomous driving applications.

\begin{table}[t]
\caption{Quantitative comparison on the NuScenes dataset for scene reconstruction and novel view synthesis tasks.}
\label{tab:nuscenes}
\centering
\scriptsize
\setlength{\tabcolsep}{4pt}
\renewcommand{\arraystretch}{1.12}
\begin{adjustbox}{width=0.48\textwidth} % half-column width
\begin{tabular}{lccc ccc}
\toprule
\multirow{2}{*}{Method} &
\multicolumn{3}{c}{Reconstruction} &
\multicolumn{3}{c}{Novel View} \\
\cmidrule(lr){2-4}\cmidrule(lr){5-7}
& PSNR$\uparrow$ & SSIM$\uparrow$ & LPIPS$\downarrow$
& PSNR$\uparrow$ & SSIM$\uparrow$ & LPIPS$\downarrow$ \\
\midrule
3DGS~\cite{kerbl20233d}
& 27.6  & 0.741 & 0.378
& 26.4 & 0.704 & 0.382 \\

StreetGaussians~\cite{yan2024street}
& 28.3 & 0.749 & 0.364
& 26.9 & \second{0.709} & 0.376 \\

OmniRe~\cite{chen2025omnire}
& \second{28.7} & \second{0.760} & \second{0.335}
& \second{26.9} & 0.705 & \second{0.351} \\

Ours
& \best{29.7} & \best{0.775} & \best{0.319}
& \best{27.7} & \best{0.718} & \best{0.338} \\
\bottomrule
\end{tabular}
\end{adjustbox}
\end{table}

\begin{figure*}[!t]
  \centering
  \includegraphics[width=0.92\textwidth,height=0.9\textheight,keepaspectratio]{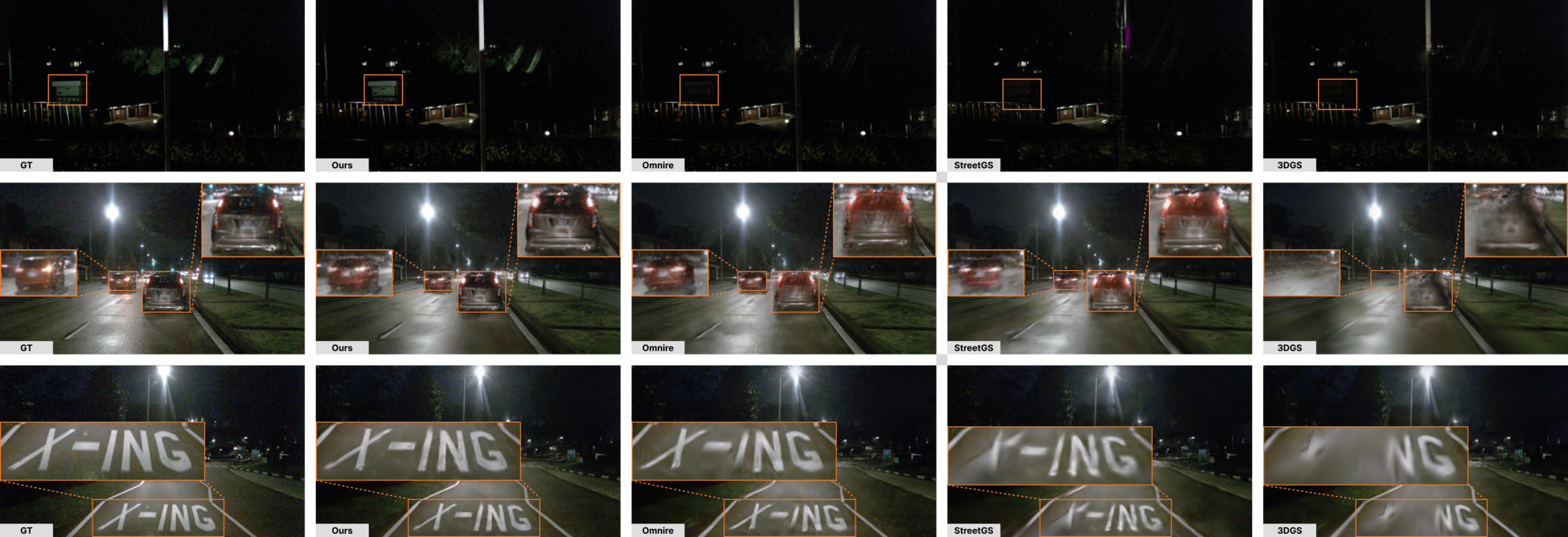}
  \caption{Qualitative results from the nuScenes Dataset. Columns show results from GT, Ours, OmniRe, StreetGS, and 3DGS, respectively. Fine-grained details such as traffic lights, vehicles, and trees are highlighted to illustrate differences in reconstruction quality.}
  \label{fig:nuScenes}
\end{figure*}

\begin{figure*}[t]
  \centering
  \includegraphics[width=0.85\linewidth,keepaspectratio]{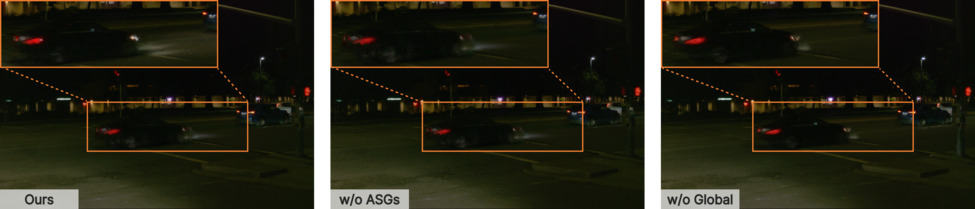}
  \caption{Comparison of our decomposed lighting modules. Without ASGs, fine details on vehicles and scene elements are lost due to oversmoothing. Without global SH illumination, the model fails to capture the interaction between light sources (e.g., vehicle headlights) and the surrounding environment, including shadows cast on the ground.}
  \label{fig:lighting}
\end{figure*}

\subsection{Results on nuScenes}

The quantitative results on nuScenes are presented in~\Cref{tab:nuscenes}. Despite the lower resolution, our model exhibits robust performance across all three metrics for both novel view synthesis and scene reconstruction.
Specifically, our framework achieves a PSNR of 29.7 dB, SSIM of 0.775, and LPIPS of 0.319 for scene reconstruction, outperforming all prior state-of-the-art frameworks. Compared to the second-best baseline, OmniRe, this corresponds to a approximately 1.0 dB improvement in PSNR and a notable reduction in LPIPS, indicating sharper and robust reconstructions.
For novel view synthesis, our model also outperforms the state-of-the-art models, achieving a PSNR of 27.7 dB, SSIM of 0.718, and LPIPS of 0.338. Although the improvements are slightly less than those for scene reconstruction, our method consistently surpasses other methods, highlighting its ability to generalize across challenging low-light driving scenes.

The qualitative results in~\Cref{fig:nuScenes} on the nuScenes dataset further demonstrate the advantages of our approach.
From~\Cref{fig:nuScenes}, we can see that our method produces clearer vehicle renderings and accurately reconstructs background elements, such as buildings, where other methods struggle. Additionally, our approach effectively reconstructs road lane signs, which are important for navigation in low-light conditions, while other methods fail to recover these features. These highlight the superiority of our framework in modeling low-light driving scenes and its ability to preserve critical details for autonomous driving tasks. 
These results confirm that our decomposition generalizes well across various scenes from different datasets, ensuring the fidelity of the our framework.

\begin{table}[t]
\caption{Ablation study on the effectiveness of our decomposed lighting mechanism for scene reconstruction and novel view synthesis tasks.}
\label{tab:ablation1}
\centering
\scriptsize
\setlength{\tabcolsep}{3pt}
\renewcommand{\arraystretch}{1.12}
\begin{adjustbox}{width=0.48\textwidth} % half-column width
\begin{tabular}{lccc ccc}
\toprule
\multirow{2}{*}{Variant} &
\multicolumn{3}{c}{Reconstruction} &
\multicolumn{3}{c}{Novel View}  \\
\cmidrule(lr){2-4}\cmidrule(lr){5-7}
& PSNR$\uparrow$ & SSIM$\uparrow$ & LPIPS$\downarrow$ 
& PSNR$\uparrow$ & SSIM$\uparrow$ & LPIPS$\downarrow$ \\
\midrule
w/o Global Diffuse Modeling
& 30.1 & 0.765 & 0.397
& 27.9 & 0.716 & 0.417 
\\

w/o Specular Modeling
& 29.8 & 0.767 & 0.390 
& 27.9 & 0.713 & 0.416 
\\

Full Model
& \best{30.8} & \best{0.778} & \best{0.380} 
& \best{28.2} & \best{0.720} & \best{0.403} 
\\
\bottomrule
\end{tabular}
\end{adjustbox}
\end{table}

\subsection{Ablation Study}

\subsubsection{Effectiveness of Lighting Decomposition}
As shown in~\Cref{tab:ablation1}, removing either the global diffuse SH-based illumination or the per-Gaussian ASG-based lighting degrades performance in both novel view and reconstruction tasks. Without global illumination, the model struggles to maintain consistency across the scene, leading to reduced PSNR and SSIM. Without ASG-based specular modeling, localized highlights such as vehicle headlights and reflective building windows are lost, causing a drop in perceptual quality (higher LPIPS). 
Furthermore, as shown in~\Cref{fig:lighting}, the module without ASGs fails to render fine details and oversmooths the vehicles, while removing the global SH illumination results in incorrect modeling of the interaction between vehicle headlights and the ground and the shadows cast by the lighting.
These results confirm that both global and local lighting are complementary and necessary for robust rendering.

\begin{table}[t]
\caption{Ablation study on the effectiveness of different specular modeling techniques.}
\label{tab:ablation2}
\centering
\scriptsize
\setlength{\tabcolsep}{3pt}
\renewcommand{\arraystretch}{1.12}
\begin{adjustbox}{width=0.48\textwidth} % half-column width
\begin{tabular}{lccc ccc}
\toprule
\multirow{2}{*}{Variant} &
\multicolumn{3}{c}{Reconstruction} &
\multicolumn{3}{c}{Novel View} \\
\cmidrule(lr){2-4}\cmidrule(lr){5-7}
& PSNR$\uparrow$ & SSIM$\uparrow$ & LPIPS$\downarrow$ 
& PSNR$\uparrow$ & SSIM$\uparrow$ & LPIPS$\downarrow$ \\
\midrule

Specular Modeling w/ SH 
& 30.1 & 0.768 & 0.391
& 28.1 & 0.717 & 0.407 \\

Specular Modeling w/ ASG 
& \best{30.8} & \best{0.778} & \best{0.380}
& \best{28.2} & \best{0.720} & \best{0.403} \\
\bottomrule
\end{tabular}
\end{adjustbox}
\end{table}

\begin{figure}[t]
  \centering
  \includegraphics[width=0.9\columnwidth]{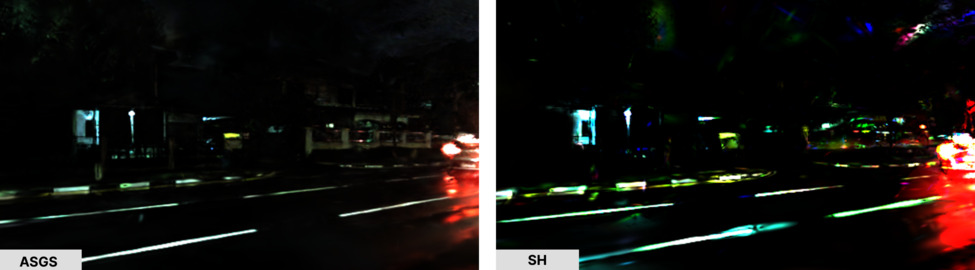}
  \caption{Comparison of specular maps produced using SH and ASGs. SH-based incident specular lighting tends to create exaggerated highlights that deviate from realistic appearance, whereas ASGs capture sharper and more physically plausible specular effects.}
  \label{fig:sh_asgs}
\end{figure}

\subsubsection{Analysis of ASG vs. SH for Specular Modeling}
As shown in~\Cref{tab:ablation2}, replacing ASGs with SH for incident lighting yields similar PSNR and SSIM, but increases in LPIPS, reflecting poorer perceptual sharpness.
Qualitatively, as shown in~\Cref{fig:sh_asgs}, we observe that SH fails to capture sharp specular highlights and instead produces smoother, overly diffuse reflections. ASGs, with their anisotropic nature, more accurately reproduce high-frequency specular effects, making them better suited for specular modeling.

\begin{table}[t]
\caption{Ablation study on the effectiveness of BRDF Constraint.}
\label{tab:ablation3}
\centering
\scriptsize
\setlength{\tabcolsep}{3pt}
\renewcommand{\arraystretch}{1.12}
\begin{adjustbox}{width=0.48\textwidth} % half-column width
\begin{tabular}{lccc ccc}
\toprule
\multirow{2}{*}{Variant} &
\multicolumn{3}{c}{Reconstruction} &
\multicolumn{3}{c}{Novel View} \\
\cmidrule(lr){2-4}\cmidrule(lr){5-7}
& PSNR$\uparrow$ & SSIM$\uparrow$ & LPIPS$\downarrow$ 
& PSNR$\uparrow$ & SSIM$\uparrow$ & LPIPS$\downarrow$ \\
\midrule

Ours w/o BRDF Constraint
& 29.8 & 0.760 & 0.426
& 27.6 & 0.712 & 0.446 \\

Ours w/ BRDF Constraint
& \best{30.8} & \best{0.778} & \best{0.380}
& \best{28.2} & \best{0.720} & \best{0.403} \\
\bottomrule
\end{tabular}
\end{adjustbox}
\end{table}

\begin{figure}[t]
  \centering
  \includegraphics[width=0.9\columnwidth]{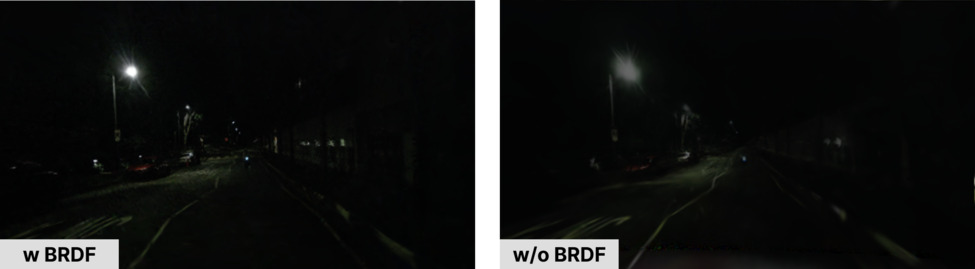}
  \caption{Comparison of BRDF-constrained and direct evaluation of ASGs. BRDF-constrained ASGs produces more physically realistic lighting and construct overall better quality rendering than direct evaluation.}
  \label{fig:brdf}
\end{figure}

\subsubsection{Analysis of BRDF Constraint}
As shown in~\Cref{tab:ablation3}, when ASGs are directly evaluated without BRDF constraints, the performance degrades dramatically in all metrics compared to BRDF-constraint ASGs. Specifically, PSNR drops by 1.0 dB with a significant increase in LPIPS. 
This aligns with the result reported by~\cite{yang2024spec}.
Furthermore, a comparison of BRDF-constrained and direct evaluation of ASGs is shown in~\Cref{fig:brdf}.
This confirms that constraining specular component with a physically based BRDF framework is crucial for producing realistic and physically consistent lighting, rather than arbitrary highlight patterns as in~\Cref{fig:brdf}.

\begin{figure}[!h]
  \centering
  \includegraphics[width=0.9\columnwidth]{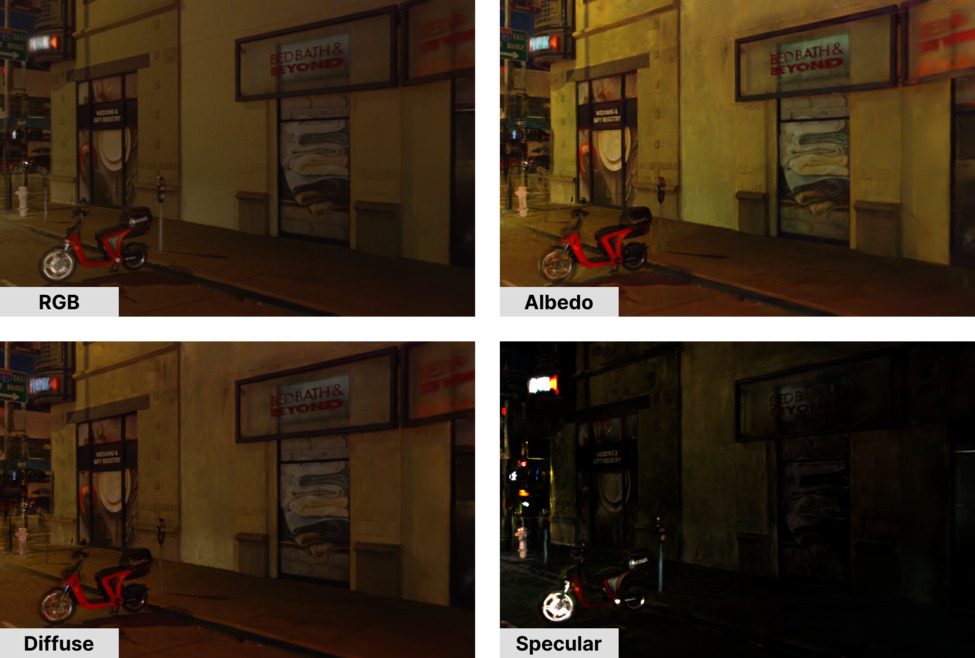}
  \caption{Visualization of the outputs. From top to bottom and left to right:  the final rendered image, albedo, diffuse, and specular. This illustrates the decomposition and the contribution of each lighting component to the final rendering.}
  \label{fig:result}
\end{figure}

\subsubsection{A Visualization of Lighting Component Decomposition.}
We show a visualization of lighting component decomposition in~\Cref{fig:result}.
We can see that the albedo represents the base color of the scene without any lighting or shading.
The diffuse component models non-directional light and accounts for shading caused by environmental illumination, while the specular component captures sharp highlights from reflective surfaces or self-illuminating objects, such as neon signs or traffic signals. 
These results demonstrate the robustness of our framework and its ability to produce physically plausible and realistic lighting decomposition.

\section{Conclusion}
This work proposes a novel framework to effectively reconstruct low-light autonomous driving scenes.
Our key idea is to integrate physically based rendering into composite scene Gaussian representations, which jointly optimizes BRDF-based material properties, for nighttime driving scene reconstruction.
Experiments on two challenging autonomous driving datasets demonstrate the superiority of our framework both quantitatively and qualitatively.

\noindent{\textbf{Limitation and Future Work.}}
Since our approach adopts a per-sequence reconstruction paradigm, our approach has scalability issues when applied to large-scale driving scene simulation in digital twin environments.
In addition, the current lighting modeling strategy may not fully capture complex lighting conditions in large-scale driving scenes.
Our future work aims to address these limitations by exploring feed-forward reconstruction paradigms, incorporating richer priors beyond surface normals to better capture material properties, exploring alternative BRDF models other than the simplified Disney model, and integrating more advanced tone-mapping module to further enhance reconstruction quality.

\bibliographystyle{IEEEtran}  
\bibliography{ref}

@inproceedings{chen2025omnire,
  title={OmniRe: Omni Urban Scene Reconstruction},
  author={Ziyu Chen and Jiawei Yang and Jiahui Huang and Riccardo de Lutio and Janick Martinez Esturo and Boris Ivanovic and Or Litany and Zan Gojcic and Sanja Fidler and Marco Pavone and Li Song and Yue Wang},
  booktitle={The Thirteenth International Conference on Learning Representations},
  year={2025}
}

@inproceedings{ost2021neural,
  title={Neural scene graphs for dynamic scenes},
  author={Ost, Julian and Mannan, Fahim and Thuerey, Nils and Knodt, Julian and Heide, Felix},
  booktitle={Proceedings of the IEEE/CVF Conference on Computer Vision and Pattern Recognition},
  pages={2856--2865},
  year={2021}
}

@article{yang2023emernerf,
  title={Emernerf: Emergent spatial-temporal scene decomposition via self-supervision},
  author={Yang, Jiawei and Ivanovic, Boris and Litany, Or and Weng, Xinshuo and Kim, Seung Wook and Li, Boyi and Che, Tong and Xu, Danfei and Fidler, Sanja and Pavone, Marco and others},
  journal={arXiv preprint arXiv:2311.02077},
  year={2023}
}

@inproceedings{yang2023unisim,
  title={Unisim: A neural closed-loop sensor simulator},
  author={Yang, Ze and Chen, Yun and Wang, Jingkang and Manivasagam, Sivabalan and Ma, Wei-Chiu and Yang, Anqi Joyce and Urtasun, Raquel},
  booktitle={Proceedings of the IEEE/CVF Conference on Computer Vision and Pattern Recognition},
  pages={1389--1399},
  year={2023}
}

@inproceedings{gao2024relightable,
  title={Relightable 3d gaussians: Realistic point cloud relighting with brdf decomposition and ray tracing},
  author={Gao, Jian and Gu, Chun and Lin, Youtian and Li, Zhihao and Zhu, Hao and Cao, Xun and Zhang, Li and Yao, Yao},
  booktitle={European Conference on Computer Vision},
  pages={73--89},
  year={2024},
  organization={Springer}
}

@article{xu2013anisotropic,
  title={Anisotropic spherical gaussians},
  author={Xu, Kun and Sun, Wei-Lun and Dong, Zhao and Zhao, Dan-Yong and Wu, Run-Dong and Hu, Shi-Min},
  journal={ACM Transactions on Graphics (TOG)},
  volume={32},
  number={6},
  pages={1--11},
  year={2013},
  publisher={ACM New York, NY, USA}
}

@incollection{reinhard2023photographic,
  title={Photographic tone reproduction for digital images},
  author={Reinhard, Erik and Stark, Michael and Shirley, Peter and Ferwerda, James},
  booktitle={Seminal Graphics Papers: Pushing the Boundaries, Volume 2},
  pages={661--670},
  year={2023}
}

@inproceedings{ye2024gaussian,
  title={Gaussian in the Dark: Real-Time View Synthesis From Inconsistent Dark Images Using Gaussian Splatting},
  author={Ye, Sheng and Dong, Zhen-Hui and Hu, Yubin and Wen, Yu-Hui and Liu, Yong-Jin},
  booktitle={Computer Graphics Forum},
  volume={43},
  number={7},
  pages={e15213},
  year={2024},
  organization={Wiley Online Library}
}

@inproceedings{burley2012physically,
  title={Physically-based shading at disney},
  author={Burley, Brent and Studios, Walt Disney Animation},
  booktitle={Acm siggraph},
  volume={2012},
  number={2012},
  pages={1--7},
  year={2012},
  organization={vol. 2012}
}

@inproceedings{green2003spherical,
  title={Spherical harmonic lighting: The gritty details},
  author={Green, Robin},
  booktitle={Archives of the game developers conference},
  volume={56},
  pages={4},
  year={2003}
}

@inproceedings{yan2024street,
  title={Street gaussians: Modeling dynamic urban scenes with gaussian splatting},
  author={Yan, Yunzhi and Lin, Haotong and Zhou, Chenxu and Wang, Weijie and Sun, Haiyang and Zhan, Kun and Lang, Xianpeng and Zhou, Xiaowei and Peng, Sida},
  booktitle={European Conference on Computer Vision},
  pages={156--173},
  year={2024},
  organization={Springer}
}

@article{sun2025ll,
  title={LL-Gaussian: Low-Light Scene Reconstruction and Enhancement via Gaussian Splatting for Novel View Synthesis},
  author={Sun, Hao and Yu, Fenggen and Xu, Huiyao and Zhang, Tao and Zou, Changqing},
  journal={arXiv preprint arXiv:2504.10331},
  year={2025}
}

@article{yang2024spec,
  title={Spec-gaussian: Anisotropic view-dependent appearance for 3d gaussian splatting},
  author={Yang, Ziyi and Gao, Xinyu and Sun, Yang-Tian and Huang, Yihua and Lyu, Xiaoyang and Zhou, Wen and Jiao, Shaohui and Qi, Xiaojuan and Jin, Xiaogang},
  journal={Advances in Neural Information Processing Systems},
  volume={37},
  pages={61192--61216},
  year={2024}
}

@article{chen2024vcr,
  title={Vcr-gaus: View consistent depth-normal regularizer for gaussian surface reconstruction},
  author={Chen, Hanlin and Wei, Fangyin and Li, Chen and Huang, Tianxin and Wang, Yunsong and Lee, Gim Hee},
  journal={Advances in Neural Information Processing Systems},
  volume={37},
  pages={139725--139750},
  year={2024}
}

@article{wang2025moge,
  title={MoGe-2: Accurate Monocular Geometry with Metric Scale and Sharp Details},
  author={Wang, Ruicheng and Xu, Sicheng and Dong, Yue and Deng, Yu and Xiang, Jianfeng and Lv, Zelong and Sun, Guangzhong and Tong, Xin and Yang, Jiaolong},
  journal={arXiv preprint arXiv:2507.02546},
  year={2025}
}

@article{chen2023periodic,
  title={Periodic vibration gaussian: Dynamic urban scene reconstruction and real-time rendering},
  author={Chen, Yurui and Gu, Chun and Jiang, Junzhe and Zhu, Xiatian and Zhang, Li},
  journal={arXiv preprint arXiv:2311.18561},
  year={2023}
}

@inproceedings{yang2024deformable,
  title={Deformable 3d gaussians for high-fidelity monocular dynamic scene reconstruction},
  author={Yang, Ziyi and Gao, Xinyu and Zhou, Wen and Jiao, Shaohui and Zhang, Yuqing and Jin, Xiaogang},
  booktitle={Proceedings of the IEEE/CVF conference on computer vision and pattern recognition},
  pages={20331--20341},
  year={2024}
}

@article{kerbl20233d,
  title={3D Gaussian splatting for real-time radiance field rendering.},
  author={Kerbl, Bernhard and Kopanas, Georgios and Leimk{\"u}hler, Thomas and Drettakis, George},
  journal={ACM Trans. Graph.},
  volume={42},
  number={4},
  pages={139--1},
  year={2023}
}

@inproceedings{zhang2024darkgs,
  title={Darkgs: Learning neural illumination and 3d gaussians relighting for robotic exploration in the dark},
  author={Zhang, Tianyi and Huang, Kaining and Zhi, Weiming and Johnson-Roberson, Matthew},
  booktitle={2024 IEEE/RSJ International Conference on Intelligent Robots and Systems (IROS)},
  pages={12864--12871},
  year={2024},
  organization={IEEE}
}

@inproceedings{zhou2024drivinggaussian,
  title={Drivinggaussian: Composite gaussian splatting for surrounding dynamic autonomous driving scenes},
  author={Zhou, Xiaoyu and Lin, Zhiwei and Shan, Xiaojun and Wang, Yongtao and Sun, Deqing and Yang, Ming-Hsuan},
  booktitle={Proceedings of the IEEE/CVF conference on computer vision and pattern recognition},
  pages={21634--21643},
  year={2024}
}

@inproceedings{zhou2024hugs,
  title={Hugs: Holistic urban 3d scene understanding via gaussian splatting},
  author={Zhou, Hongyu and Shao, Jiahao and Xu, Lu and Bai, Dongfeng and Qiu, Weichao and Liu, Bingbing and Wang, Yue and Geiger, Andreas and Liao, Yiyi},
  booktitle={Proceedings of the IEEE/CVF Conference on Computer Vision and Pattern Recognition},
  pages={21336--21345},
  year={2024}
}

@article{wu2025armgs,
  title={ArmGS: Composite Gaussian Appearance Refinement for Modeling Dynamic Urban Environments},
  author={Wu, Guile and Bai, Dongfeng and Liu, Bingbing},
  journal={arXiv preprint arXiv:2507.03886},
  year={2025}
}

@inproceedings{yao2022neilf,
  title={Neilf: Neural incident light field for physically-based material estimation},
  author={Yao, Yao and Zhang, Jingyang and Liu, Jingbo and Qu, Yihang and Fang, Tian and McKinnon, David and Tsin, Yanghai and Quan, Long},
  booktitle={European conference on computer vision},
  pages={700--716},
  year={2022},
  organization={Springer}
}

@inproceedings{qu2024lush,
  title={LuSh-NeRF: lighting up and sharpening NeRFs for low-light scenes},
  author={Qu, Zefan and Xu, Ke and Hancke, Gerhard Petrus and Lau, Rynson WH},
  booktitle={Proceedings of the 38th International Conference on Neural Information Processing Systems},
  pages={109871--109893},
  year={2024}
}

@inproceedings{caesar2020nuscenes,
  title={nuscenes: A multimodal dataset for autonomous driving},
  author={Caesar, Holger and Bankiti, Varun and Lang, Alex H and Vora, Sourabh and Liong, Venice Erin and Xu, Qiang and Krishnan, Anush and Pan, Yu and Baldan, Giancarlo and Beijbom, Oscar},
  booktitle={Proceedings of the IEEE/CVF conference on computer vision and pattern recognition},
  pages={11621--11631},
  year={2020}
}

@inproceedings{sun2020scalability,
  title={Scalability in perception for autonomous driving: Waymo open dataset},
  author={Sun, Pei and Kretzschmar, Henrik and Dotiwalla, Xerxes and Chouard, Aurelien and Patnaik, Vijaysai and Tsui, Paul and Guo, James and Zhou, Yin and Chai, Yuning and Caine, Benjamin and others},
  booktitle={Proceedings of the IEEE/CVF conference on computer vision and pattern recognition},
  pages={2446--2454},
  year={2020}
}

@article{ren2024unigaussian,
  title={Unigaussian: Driving scene reconstruction from multiple camera models via unified gaussian representations},
  author={Ren, Yuan and Wu, Guile and Li, Runhao and Yang, Zheyuan and Liu, Yibo and Chen, Xingxin and Cao, Tongtong and Liu, Bingbing},
  journal={arXiv preprint arXiv:2411.15355},
  year={2024}
}

@inproceedings{cao2024lightning,
  title={Lightning nerf: Efficient hybrid scene representation for autonomous driving},
  author={Cao, Junyi and Li, Zhichao and Wang, Naiyan and Ma, Chao},
  booktitle={2024 IEEE International Conference on Robotics and Automation (ICRA)},
  pages={16803--16809},
  year={2024},
  organization={IEEE}
}

@book{cohen1993radiosity,
  title={Radiosity and realistic image synthesis},
  author={Cohen, Michael F and Wallace, John R},
  year={1993},
  publisher={Morgan Kaufmann}
}

@inproceedings{zhang2018unreasonable,
  title={The unreasonable effectiveness of deep features as a perceptual metric},
  author={Zhang, Richard and Isola, Phillip and Efros, Alexei A and Shechtman, Eli and Wang, Oliver},
  booktitle={Proceedings of the IEEE conference on computer vision and pattern recognition},
  pages={586--595},
  year={2018}
}

@article{wang2004image,
  title={Image quality assessment: from error visibility to structural similarity},
  author={Wang, Zhou and Bovik, Alan C and Sheikh, Hamid R and Simoncelli, Eero P},
  journal={IEEE transactions on image processing},
  volume={13},
  number={4},
  pages={600--612},
  year={2004},
  publisher={IEEE}
}

@inproceedings{kajiya1986rendering,
  title={The rendering equation},
  author={Kajiya, James T},
  booktitle={Proceedings of the 13th annual conference on Computer graphics and interactive techniques},
  pages={143--150},
  year={1986}
}

@article{kingma2014adam,
  title={Adam: A method for stochastic optimization},
  author={Kingma, Diederik P},
  journal={arXiv preprint arXiv:1412.6980},
  year={2014}
}

\end{document}